\documentclass{article}

 \usepackage[preprint]{nips_2018}

\usepackage[utf8]{inputenc} 
\usepackage[T1]{fontenc}    
\usepackage{hyperref}       
\usepackage{url}            
\usepackage{booktabs}       
\usepackage{amsfonts}       
\usepackage{nicefrac}       
\usepackage{microtype}      

\usepackage{mathtools}
\usepackage{graphicx}
\usepackage{natbib}
\setcitestyle{numbers, open={(},close={)}}

\title{Program Synthesis Through Reinforcement Learning Guided Tree Search}

\author{
  Riley Simmons-Edler \\
  Department of Computer Science\\
  Princeton University\\
  Princeton, NJ 08540 \\
  \texttt{rileys@cs.princeton.edu} \\
  \And
  Anders Miltner \\
  Department of Computer Science\\
  Princeton University\\
  Princeton, NJ 08540 \\
  \texttt{amiltner@cs.princeton.edu} \\
  \And
  Sebastian Seung \\
  Department of Computer Science\\
  Princeton University\\
  Princeton, NJ 08540 \\
  \texttt{sseung@princeton.edu} \\
}

\begin{document}

\maketitle

\begin{abstract}
\emph{Program Synthesis} is the task of generating a program from a provided specification. Traditionally, this has been treated as a search problem by the programming languages (PL) community and more recently as a supervised learning problem by the machine learning community. Here, we propose a third approach, representing the task of synthesizing a given program as a Markov decision process solvable via reinforcement learning(RL). From observations about the states of partial programs, we attempt to find a program that is optimal over a provided reward metric on pairs of programs and states. We instantiate this approach on a subset of the RISC-V assembly language operating on floating point numbers, and as an optimization inspired by search-based techniques from the PL community, we combine RL with a priority search tree.
We evaluate this instantiation and demonstrate the effectiveness of our combined method compared to a variety of baselines, including a pure RL ablation and a state of the art Markov chain Monte Carlo search method on this task.
\end{abstract}

\section{Introduction}
\label{sec:intro}
Within both the programming languages and machine learning
communities, there has been a renaissance in \emph{program synthesis},
the task of automatically generating computer code from a
user-provided specification. Due to increased computational power and
the rise of deep neural networks, this once intractable problem has
now become realizable, and has already seen use in many domains~\cite{synthesissurvey},
including improving programmer efficiency by automating routine
tasks~\cite{morpheus} providing an intuitive interface for non-experts
without programming knowledge~\cite{gulwani-popl-2014}, and reducing bugs and
improving runtime efficiency for performance-critical code~\cite{stoke}.

Within the programming languages(PL) community, program synthesis is typically
solved using \emph{enumerative search} -- finding correct programs for a given
specification by na\"ively enumerating candidates until a satisfying program is
found~\cite{feser-pldi-2015,frankle+:popl16,escher}. This approach is made
tractable by narrowing the search space through integrating deductive components
into the search
process~\cite{katayama-pepm-2012,kitzelmann-thesis-2010,le-pldi-2014}, or by
modifying the language of interest into an equivalent language with a narrower
search space, and searching within that
space~\cite{optician,gvero-pldi-2013,atp}. Another common approach is to reduce
synthesis tasks to finding a satisfying assignment to a Boolean formula via a
SAT solver~\cite{combinatorial-sketching,cegun}, but this merely pushes the enumerative search into the solver.

Researchers in the machine learning(ML) community have attacked this problem
from a different direction -- rather than searching na\"ively through a restricted
space, correct programs can be efficiently found within a larger search space by
intelligently searching or sampling from that space using a learned model of how
specifications map to programs. Most of these methods use a supervised learning
approach -- training on a large synthetic dataset of input/output examples and
corresponding satisfying programs, then using recurrent neural
network(RNN)-based models to output each line in the target program successively
given a corresponding set of input/output
examples~\cite{robustfill,ap,deepcoder,neural-compilation,terpret}. Conceptually, this
approach should compose well with PL techniques -- intelligently searching through a
small search space is much easier than intelligently searching through a large
search space. Why then are these techniques not commonly used together?

Most existing work in the ML community relies heavily on the structure
of the domain specific language(DSL) being synthesized to achieve good
performance, and can't generalize to new languages easily.  These
methods often require language features such as full differentiability
of the language~\cite{neural-compilation,terpret} or a large training set of specification and
satisfying program pairs~\cite{robustfill,ap,deepcoder}.  These requirements make it difficult to
combine existing ML-based methods with techniques developed by the PL
community, which require very different properties in a DSL for
effective synthesis.  Further, most existing methods have heavily focused
on solving programs in a ``one-shot'' fashion, either successfully
outputting a satisfying program for a given specification in a single or small number of attempts or failing to do so, which scales poorly as program spaces become larger.  In contrast, as search-based methods
enumerate all programs, they can solve any synthesis task eventually,
but they may take infeasible amounts of time to do so.

\begin{figure}
	\begin{center}
		\includegraphics[width=\linewidth]{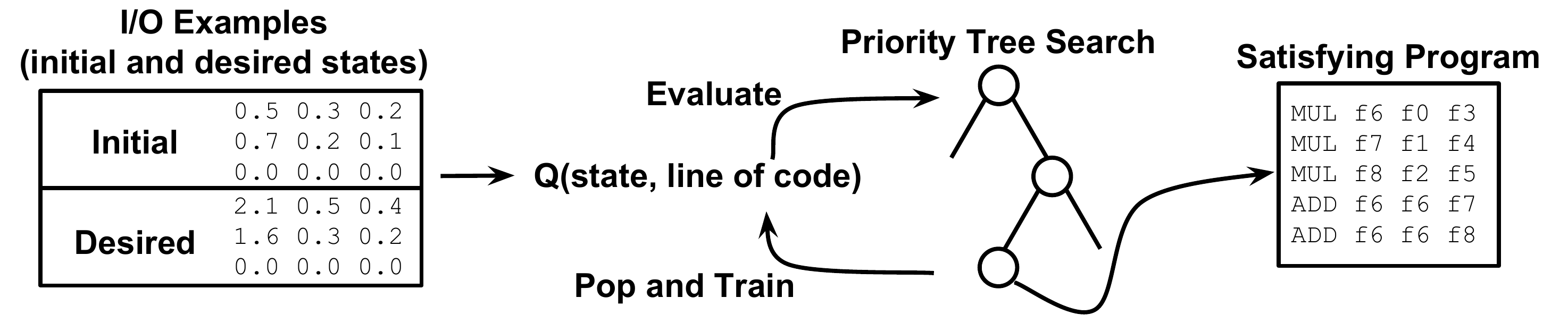}
	\end{center}
    \caption{Schematic of our proposed system. Our approach to program synthesis treats the problem as a Markov decision process solvable by reinforcement learning, and combines this with a prioritized search tree to speed up the solving process by avoiding local minima, improving the number of programs solvable within a fixed number of attempts. Given a set of input memory states and corresponding output memory states, this approach seeks to learn a policy which outputs a sequence of lines of code that maps each input example to the corresponding output example, using a reward function defined for partially-correct solutions to guide the learning process.}
    \label{fig:figure1}
\end{figure}

\textbf{Reinforcement Learning Guided Tree Search}

Our approach, \emph{reinforcement learning guided tree search (RLGTS)}, illustrated in figure \ref{fig:figure1}, seeks to combine the benefits of both search-based and
ML-based methods, and allow for the combination of techniques from
both research communities to further enhance performance. We propose a
new approach for program synthesis, representing the process of
synthesizing a program as a Markov Decision Process(MDP)\cite{rl_introduction} and
using reinforcement learning(RL) to learn to solve a program given
only a set of input/output examples for that program, a language specification,
and a reward function for the \emph{quality} of a given program. In our
RL-based approach, we interpret the program state and current partial program as an environment,
and lines of code in the language as actions seeking to maximize the reward
function. Furthermore, we combine our RL model with a
tree-based search technique which dramatically improves the performance of the method. This combination helps address issues of local minima and efficient sampling which
arise in many RL applications.

RLGTS does not depend on the availability of training data for
a given language, and makes no assumptions about the structure of the
language other than that the language allows for partial programs to
be executed and evaluated.  Further, our RL-based approach can be
combined with other program synthesis methods easily and naturally,
allowing for users to benefit from the extensive work on search-based
synthesis available for some domains.

\textbf{In summary, we make the following contributions:}
\begin{enumerate}
\item We introduce reinforcement learning guided tree search, an approach to program synthesis
that interprets program generation as a reinforcement learning task.
\item We describe an implementation of RLGTS on a subset of the RISC-V assembly
  language, and created an RL agent for this task by combining a
  Q-network-based policy with a simple search tree method.
\item We demonstrate improvements in the fraction of programs solved of up to 100\% and 800\% compared to RL-only
  and enumerative search-only baselines respectively on a synthetic dataset of random programs.  Furthermore, we compare RLGTS to
  a Markov chain Monte Carlo(MCMC) based method that has been used to great success in 
  super-optimizing x86 code and represents the current state of the art for synthesizing assembly language code~\cite{stoke}, and show superior performance on more challenging benchmark programs, solving up to 400\% more programs within a fixed program evaluation limit and remaining competitive in total performance even when that limit is increased 50x for MCMC.
\end{enumerate}

\section{Methods}
\label{sec:methods}

\subsection{Reinforcement Learning Model of Synthesis}
\label{subsec:rl_notation}

Here we describe our formulation of the general program synthesis task
as a multi-step Markov decision process, solvable via reinforcement
learning.

In the standard terminology~\cite{rl_introduction}, a fully-observed MDP is a
process having some state $s_t$ for timesteps
$t\in\{1,2,\dots,T\}$. At each state $s_t$ an action $a_t{\in}\{1,
  \dots,n\}$ from among $n$ possible actions is emitted, with some
unknown function $p(s_{t+1}|s_t, a_t)$ determining the following state
$s_{t+1}$ from among some (typically large) state space. Actions $a_t$
are selected by an agent $A(a_t|s_t; \theta)$ with a policy for selecting actions parametrized by learned parameters
$\theta$, commonly a neural network. This policy is trained to maximize
the expected cumulative reward value $E(R_T)$ emitted by some reward
function $r(s_t,a_t)$, with $R_T = \sum_{t=1}^{T}
\gamma^tr(s_t,a_t)$, with time decay factor $\gamma$.

Based on these definitions, we represent program synthesis as
follows. A \emph{program} of length $T$ is a set of actions $P_T =
\{a_1,a_2,\dots,a_T\}$. Each action $a_{t{\in}\{1,\dots,T\}}$ represents a single line of
code(e.g. "ADD f0 f1 f2") applied to state $s_t$, which represents the
memory state(the values of all variables) after execution of all
previous lines of code $\{a_1,\dots,a_{t-1}\}$ applied to a set of initial
variable assignments. The agent's task, then, is to output the next
line of code $a_t$ given $s_t$ such
that the final program $P_T$ will maximize a user-provided cumulative reward function $R_T$ on a given set of input and output
examples $I_j$ and $O_j$ for a small number of examples $j$. To fully specify the desired behavior of the program and avoid degenerate solutions that satisfy $R$ but not the programmer's intent, we use multiple input output state pairs, typically 5.

In our instantiation on RISC-V, we allow for multiple input/output examples,
so the program state is a tuple consisting of the state of multiple executions.
Thusly, our initial state is a tuple comprised of all the input states of the examples, and our desired 
output state is a tuple consisting of all the output states of the examples.
We use a reward function combining correctness (distance from current state to
the output state) and program length as a metric for computational
complexity, but this
could easily be extended to include terms optimizing for properties
like power consumption, memory usage, network/filesystem IO, or any
other desired attribute.

It is worth noting that while this formulation does not by default
include non-linear programs containing control flow, it can theoretically be extended to support them, as well as programs containing other elements of modern programming languages
as the MDP representation of a process is Turing complete~\cite{rl_introduction},
though we leave practical exploration of these topics to further research.

\subsection{Reward Function}
\label{subsec:cost_function}

For our reward function $r(s_t,a_t)$, given a set of input-output state
pairs $(I,O)$ with $N$ pairs, we use two components. First, a metric for the
correctness of the next state for the state $s_{t+1}$ produced from each example $I_j$ compared to the target output state $O_j$, measured as
\begin{equation}
  r_{\text{correctness}}(s_t,a_t) =
  \frac{\lambda_{\text{correctness}}}{NM}\sum_{j=0}^N\sum_{k=0}^M{\frac{|O[j][k] -
      s_{t+1}[j][k]|}{|O[j][k]|}}
  \label{eq:correctness}
\end{equation}
with $M$ as the number of variables used in this program, and $\lambda_{\text{correctness}}$ as a hyperparameter weight on this component of the reward function. We express correctness
as a fraction of $O[j][k]$ to normalize across output values of
different magnitudes. To this we add a term penalizing program length
to encourage the agent to learn shorter programs as an
approximation of program computational efficiency,
\begin{equation}
  r_{\text{efficiency}}(s_t,a_t) = |P_{t}|+1
  \label{eq:performance}
\end{equation}
where $|P_t|$ is the length of the program before taking action $a_t$. Because the $r_{\text{correctness}}(s_t,a_t)$ term can become large if
$s_{t+1}$ is far from $O$, we combine these terms and scale the
resulting values such that large values of $r_{\text{correctness}}(s_t,a_t)$ become close to 0 as
\begin{equation}
  r(s_t,a_t) = \frac{\lambda_{\text{scale}}}{r_{\text{correctness}}(s_t,a_t) +
    r_{\text{efficiency}}(s_t,a_t)}
  \label{eq:reward_func}
\end{equation}
with $\lambda_{\text{scale}}$ a hyperparameter weight controlling the scaling of the reward values.
We define the space of
$(P_t,r(s_t,a_t))$ pairs for a given set of
$(I,O,a{\in}\{1,...,n\})$ as a \emph{program space}, the
discrete reward landscape which our RL agent seeks to maximize.

Our formulation of program synthesis as an MDP is quite general
and can be applied for many different reward functions provided by the
user, such as that described above. However, we make a key assumption
about the reward function, which is the presence of a (possibly sparse
and non-convex) \emph{gradient} in the reward function pointing towards
the ground truth program which the RL agent can learn and
stochastically descend. While no general guarantee can be made, and
indeed it is easy to construct reward functions which provide no
gradient,\footnote{for example, $ r(s_t,a_t) =
  \begin{cases}
    1 & \text{if } s_{t+1} = s_{o}\\
    0 & \text{otherwise}
  \end{cases}
  $} there is evidence for the existence of this gradient for
practical program domains and cost functions used in previous work~\cite{stoke,synapse}.
We also demonstrate empirically in our experiments that this gradient is
present for many short floating point arithmetic programs in RISC-V.

\subsection{RL Model}
\label{subsec:rl_model}

\begin{figure}[t]
	\begin{center}
		\includegraphics[width=0.75\linewidth]{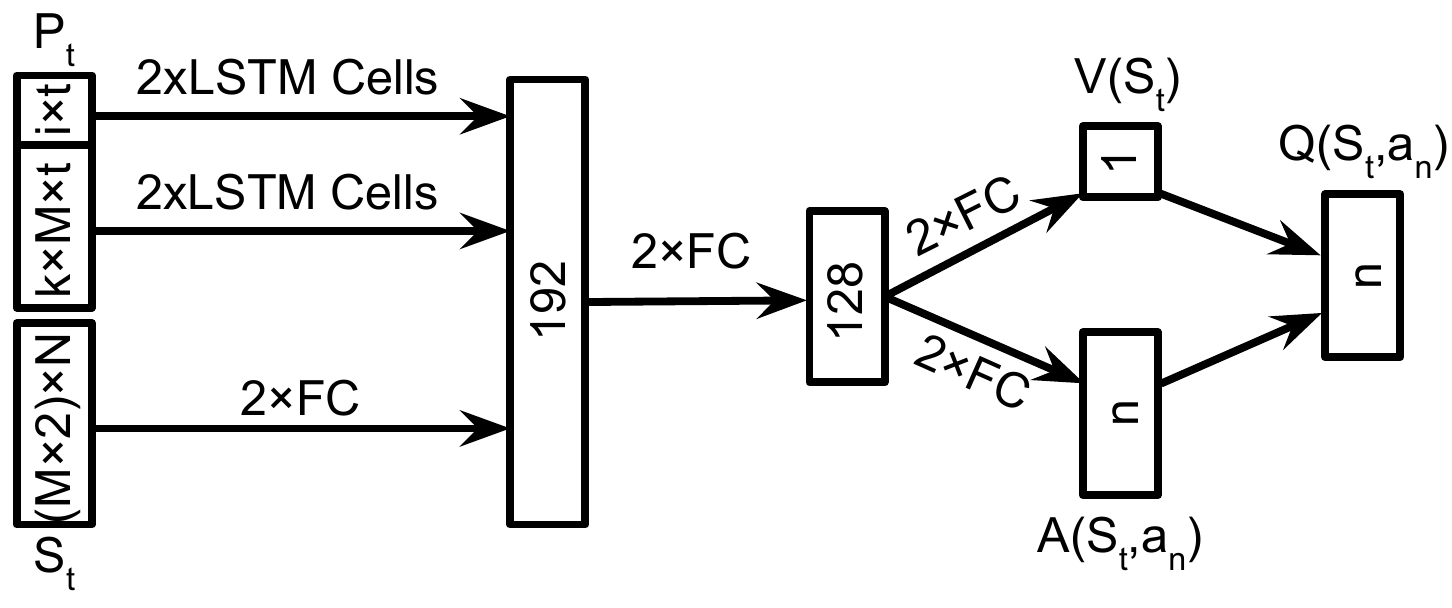}
	\end{center}
    \caption{Schematic of our Q-function neural network. We input the current memory state $S_t$ produced by a program $P_t$ of length $t$ as a set of current and target values for each of $M$ variables in use for $N$ different examples. The program that produced this memory state is encoded using an embedding for the variables of shape $k\times M\times t$ which is 1-hot in $M$ and for instructions of shape $i\times t$ 1-hot in $i$ given variables passed per instruction $k$ and number of instructions allowed $i$. These inputs are then processed through the network and used to compute the function $Q_{\theta}(S_t, a_n)$ using the dueling Q-networks formulation described in Wang et al.~\cite{dueling_q}, with $n$ the total number of actions defined by all valid combinations of instruction and variables.}
    \label{fig:agent_schematic}
\end{figure}

For reinforcement learning, we use a standard dueling double
Q-learning algorithm, which learns to predict future rewards for a
given state and action using the loss function
\begin{equation}
  \delta_t=\|r(s_t,a_t) +
  {\gamma}Q_{\theta'}(s_{t+1},\text{argmax}_{a'}Q_\theta(s_{t+1},a')) -
  Q_{\theta}(s_t, a_t)\|^2
  \label{eq:rl_objective}
\end{equation}
given agent Q-function $Q$ with parameters $\theta$ and target
Q-function parameters $\theta'$, as per Hasselt et al~\cite{double_Q}. We set the
decay term on future rewards $\gamma$ to 0.99. The objective of the Q
function trained using equation \ref{eq:rl_objective} is to predict
the expected future reward $E(R_t(s_t,a))$ that will result from
taking each action $a{\in}\{1,\dots,n\}$ possible at state
$s_t$~\cite{rl_introduction}. During synthesis, we then use an $\epsilon$-greedy policy of either taking action $\text{argmax}_a(Q_\theta(s_t,a))$ or
else taking a random action with probability $\epsilon = 0.1$.

The input to the agent consists of two parts, the state $s_t$ encoding current and target values for each variable and a sequence of
1-hot vectors encoding previous lines of the program that produced
$s_t$, $P_t$. A schematic diagram
of our network architecture is shown in Figure
\ref{fig:agent_schematic}. It consists of two input modules, for
$s_t$ and $P_t$. The
$s_t$ module is a two-layer fully connected(FC) neural
network, while the $P_t$ module is a two-layer
LSTM\cite{lstm} operating on input sequences of up to length
$p$, the maximum program length.
These modules are concatenated and fed into two more FC layers,
followed by a two-layer value stream computing $V_{\theta}(s_t)$ the expected
future reward value from being in the current state, and a separate two-layer advantage
stream computing $A_{\theta}(s_t, a_n)$ the advantage of each action
$a{\in}\{1,\dots,n\}$ above or below $V_{\theta}(s_t)$. These
modules are combined as per Wang et al.~\cite{dueling_q} to compute a Q-value per
action as 
\begin{equation}
Q_{\theta}(s_t, a_n)=V_{\theta}(s_t) + A_{\theta}(s_t,a_n) -
\frac{1}{|a_n|}\sum_{j=0}^{|n|}A_{\theta}(s_t,a_j)
\label{eq:dueling_q}
\end{equation}
with each action
$a_{j{\in}\{1,\dots,n\}}$ representing a single line of code
expressible in the language. ReLU non-linearities were used
in the FC layers, with LSTM non-linearities following the standard
arrangement of sigmoid and tanh functions~\cite{lstm}.

\subsection{Tree Search}
\label{subsec:exploration}

\begin{figure}[t]
	\begin{center}
		\includegraphics[width=\linewidth]{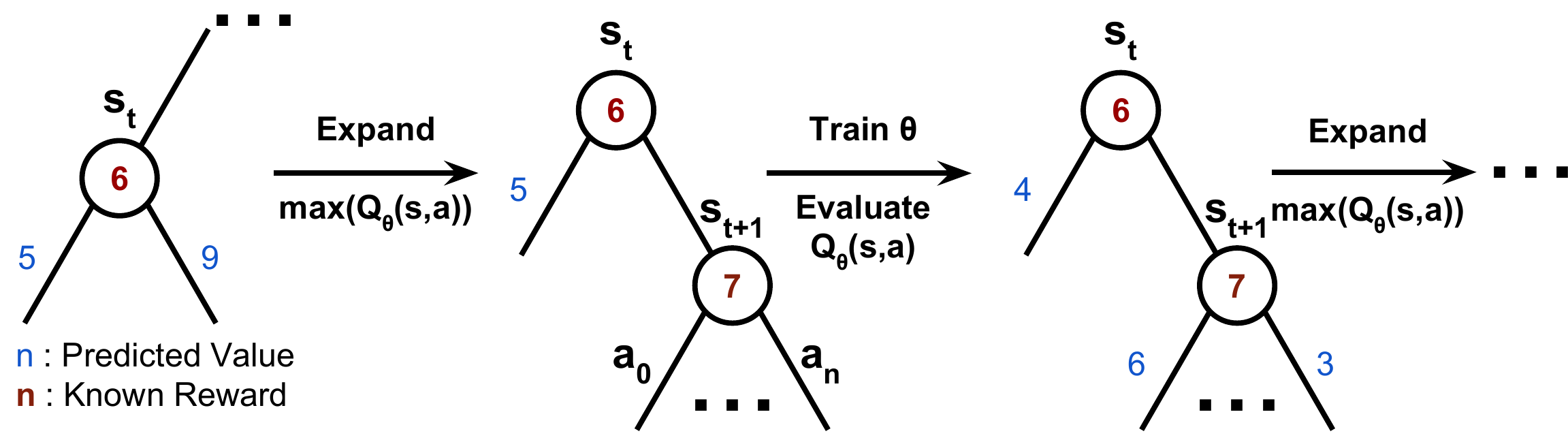}
	\end{center}
    \caption{Schematic showing our combination of Q-function and priority search tree. The Q-function specifies the priority of each unexplored edge in the tree, which then follows an $\epsilon$-greedy sampling strategy. The accumulated experiences from expanding the tree are then used to train the Q-function to improve the prioritization of edges. To reduce the cost of rescoring unexplored edges, we update the scores every 100 training iterations.}
    \label{fig:search_tree}
\end{figure}

Notably unlike most well-studied RL applications, where an agent
capable of reliably reaching high reward states across many rollouts
of the agent is desired, in program synthesis once our agent
discovers a solution to a given program we have completed synthesis
for that program and can stop training and executing our agent. We
define ``solved'' here as a candidate program producing $r(s_t,
a_t){\geq}r_{\text{GT}}$, the reward associated with the output state and GT program length. (In the case where we lack a
known ground truth solution program we could instead halt our search
once we find a program deemed sufficiently good, or after a fixed
search time interval). In either case, our ultimate objective is to
sample a set of actions leading to a high-scoring state at least once,
with repeated samplings of the same action sequence having marginal
benefit.

In line with this objective, we combined our Q-function with a simple
prioritized search tree algorithm to aid in efficient exploration.
The task of the Q-function then becomes to
predict a Q-score for each unexplored edge of the tree, the edge
representing a possible action $a{\in}\{1,\dots,n\}$ to append to
the program stored at a node defined by an existing program $P_t$ of some length $t$. The
search algorithm then simply expands
$\text{argmax}_{s,a}(Q_{\theta}(s,a))$, the edge that the current
Q-function thinks will yield the highest reward among unexplored edges, evaluating the program $P_t + a$ and adding a new node to the
tree with additional unexplored edges. We alternate between sampling
unexplored edges(using $\epsilon$-greedy
sampling) and training the Q-network, with 100 samples followed by 100
minibatch updates of the network and a rescore of all unexplored edges
with the updated Q-network. This process is illustrated in figure
\ref{fig:search_tree}. By taking this approach, we encourage
exploration by only evaluating each unique $(s,a)$ combination once,
and guarantee our
worst-case performance to be that of enumerative search, memory
and computational resources permitting. While we do not explore it
further here, this approach also allows RL-based synthesis to be
combined easily with search-based synthesis methods using deductive
reasoning and search tree pruning for DSL-specific performance
improvements.

\section{Experiments}
\label{sec:experiments}

\begin{figure}
  \label{fig:lang}
  \begin{tabular}{rcl}
  $p$ & $::=~$ & $c;p$\\
      & $~|~$ & EOF\\
  \end{tabular}\\
  \begin{tabular}{rclllllll}
    $c$
    & $::=~$ & SQRT    $\mathit{f}$ $\mathit{f}$
    & $~|~$ & ADD      $\mathit{f}$ $\mathit{f}$ $\mathit{f}$
    & $~|~$ & SUB      $\mathit{f}$ $\mathit{f}$ $\mathit{f}$
            
    & $~|~$ & MUL      $\mathit{f}$ $\mathit{f}$ $\mathit{f}$\\
    & $~|~$ & DIV      $\mathit{f}$ $\mathit{f}$ $\mathit{f}$
    & $~|~$ & SGN      $\mathit{f}$ $\mathit{f}$ $\mathit{f}$
    & $~|~$ & SGNN     $\mathit{f}$ $\mathit{f}$ $\mathit{f}$
    & $~|~$ & SGNX     $\mathit{f}$ $\mathit{f}$ $\mathit{f}$\\
    & $~|~$ & MIN      $\mathit{f}$ $\mathit{f}$ $\mathit{f}$
    & $~|~$ & MAX      $\mathit{f}$ $\mathit{f}$ $\mathit{f}$
    & $~|~$ & EQ       $\mathit{f}$ $\mathit{f}$ $\mathit{f}$
    & $~|~$ & LT       $\mathit{f}$ $\mathit{f}$ $\mathit{f}$\\
    & $~|~$ & LTE      $\mathit{f}$ $\mathit{f}$ $\mathit{f}$
    & $~|~$ & MADD     $\mathit{f}$ $\mathit{f}$ $\mathit{f}$ $\mathit{f}$
    & $~|~$ & MSUB     $\mathit{f}$ $\mathit{f}$ $\mathit{f}$ $\mathit{f}$
    & $~|~$ & NMADD    $\mathit{f}$ $\mathit{f}$ $\mathit{f}$ $\mathit{f}$\\
    & $~|~$ & NMSUB    $\mathit{f}$ $\mathit{f}$ $\mathit{f}$ $\mathit{f}$
  \end{tabular}\\
  \begin{tabular}{rcl}
  $\mathit{f}$ & $::=~$ & f1 $~|~ \ldots ~|~$ f32 \\
  \end{tabular}\\
  \caption{Grammar describing the subset of RISC-V we aim to synthesize.}
\end{figure}

\subsection{Experiment Setup}
To analyze the performance of our instantiation of RLGTS, we synthesize programs using
a common subset of the RISC-V assembly programming
language~\cite{RISCV}. We select a core subset of instructions on
floating point values, shown in Figure~\ref{fig:lang}, which
excludes control flow, memory reads/writes, and ``magic number''
inputs for simplicity. This domain is interesting
because there are few previously published methods that can perform
better than na\"ive enumerative search other than
stochastic search using hand-tuned MCMC search to estimate the program
gradient~\cite{stoke}.

To benchmark our performance, we construct a dataset of synthetic
programs by generating random programs with specified attributes. For each program in the
evaluation set, we specify the number of lines, number of allowed
instructions, number of allowed variables, and number of examples, and
synthesize a random program to satisfy these specifications, rejecting programs which we can easily determine can be expressed in fewer lines. Our
generation process allows the use of fewer types of instructions or
fewer variables, but will always obey the length and example count
specification exactly, and each method must search the entire program
space defined by the specified constraints. We generated 100 programs for each set of specified attributes.
The search spaces we traverse have sizes ranging from $2.1\times 10^6$ possible programs (3 lines, 2 instructions, 4 variables), to $4.0\times 10^{31}$ possible programs (15 lines, 2 instructions, 4 variables).

To train our network, we
use a learning rate of
0.001, and sampled batches of 64 experiences
from an experience buffer
storing all previously observed states using proportional
prioritized experience replay as per Schaul et al.~\cite{prioritized_replay}, with $\alpha = 0.6$ and $\beta_{\text{initial}} = 0.5$, linearly annealing $\beta$ to 1.0 after 10,000 iterations. We updated the target Q network parameters by
setting $Q_{\theta'} = Q_\theta$ every 100 training iterations. We set the reward function hyperparameters $\lambda_{\text{correctness}}$ and $\lambda_{\text{scale}}$ to 5 and 100 respectively for all methods.
While our system contains a number of hyperparameters that affect performance, we performed only basic manual parameter tuning on a short hand-written test program intended to be readily solvable, and a set of 10 randomly generated length 3 programs which were not used to generate our results, with the resulting parameters used for all experiments.

Each method was allowed to
run on each program in the benchmark suite until a satisfying program was found,
20,000 programs were proposed, 16 GB of memory were consumed, or 8 hours on
one 2.4GHz Broadwell CPU and one Nvidia p100 GPU had been consumed. In our comparisons, 
we focus on sample efficiency instead of clock time, and
while proposing 20,000
programs takes considerably less time and computation using purely
search based methods running on CPU, both neural network and search-based methods could have been further optimized to increase speed.
Because of this, we do not consider wall clock time
to be definitive for any method here described and thus focus on the
number of proposed programs required to solve instead as a measure of
efficiency and the potential of the method. We expect with further optimizations and
additional hardware that significantly harder programs could be tractably solved
by RLGTS.

\begin{figure}[t]
	\begin{center}
		\includegraphics[width=0.5\linewidth]{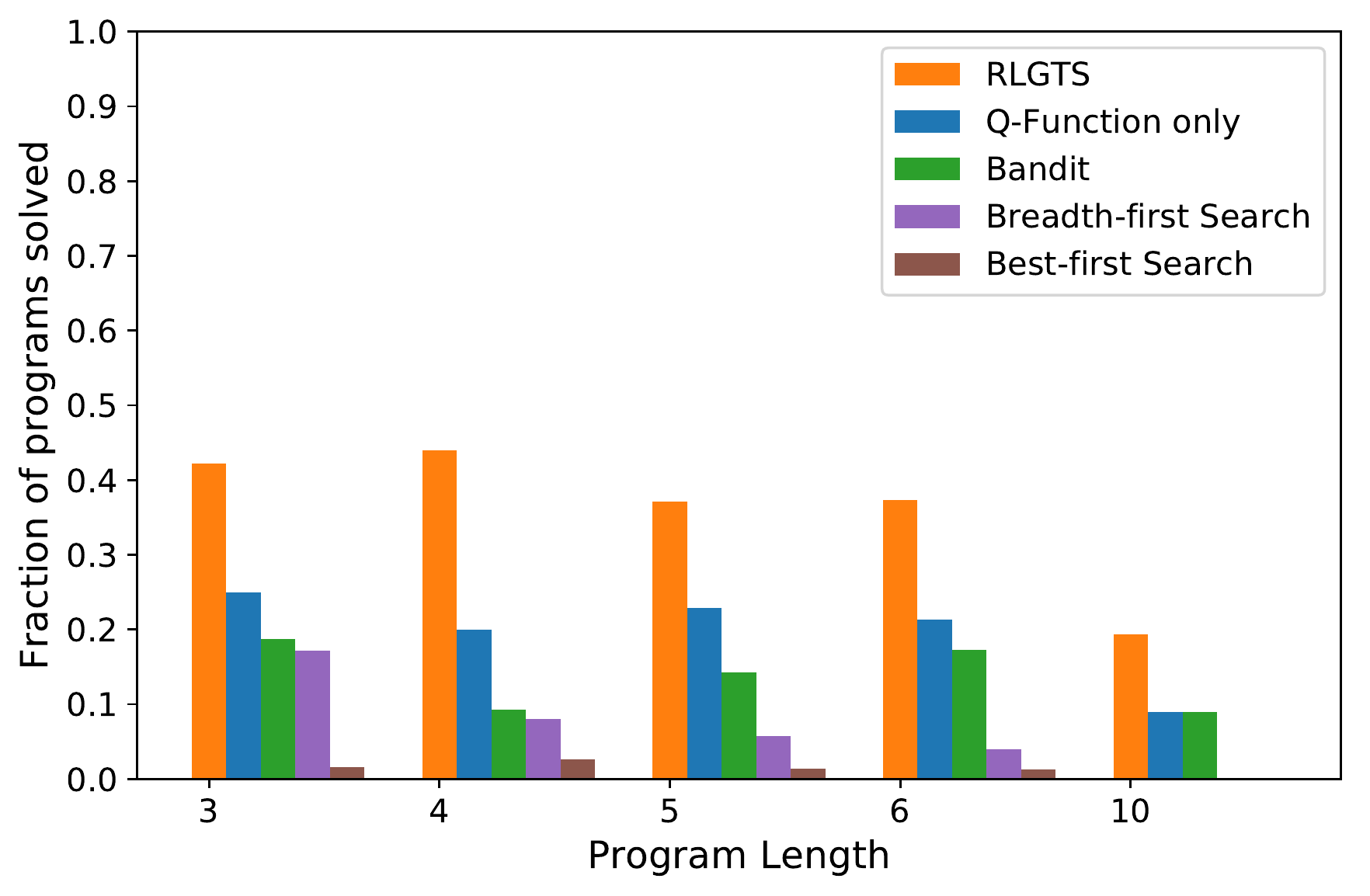}\includegraphics[width=0.5\linewidth]{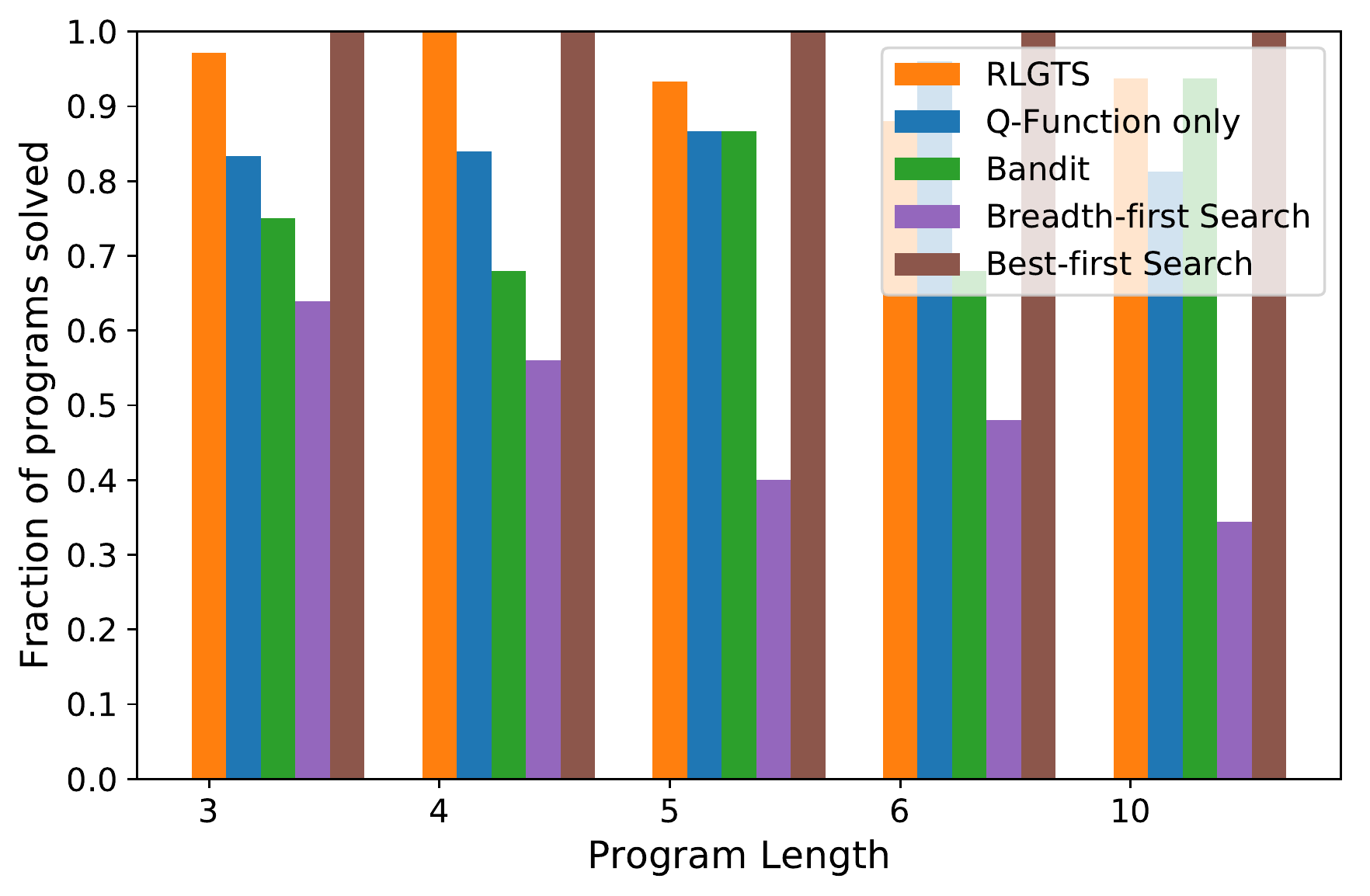}
	\end{center}
    \caption{Comparison of performance between RLGTS and baseline methods, showing fraction of non-convex programs solved in the left figure and fraction of convex programs solved in the right figure. While RLGTS is effective at solving convex programs, they can be easily solved via best-first search, and thus we remove them from our other experiments as they are not a good indicator of method performance. The maximum program length was 6 for programs of up to length 6, and 10 for programs of length 10.}
    \label{fig:prog_length}
\end{figure}

\subsection{Baselines}
\label{subsec:baselines}

We compare RLGTS to several baselines. In the simplest case, we
run a breadth-first enumerative search algorithm, which has expected
program solve time of approximately $N/2$ for a program space with $N$ possible
programs of length equal to the GT program.

Second, we compare to a simple multi-armed bandit model~\cite{bandit}
as the simplest form of RL-trainable model, representing each decision
$x$ defining a program as an independent random variable sampled based on a
learned probability $p_{\theta}(x)$, 
trained using the REINFORCE algorithm~\cite{reinforce}.

Next, we compare to a Q-learning baseline as an ablation of our full
system, using the same network architecture and training procedure,
but lacking the search tree of the full system.

Lastly, we compare to the approach used in Stoke, a heuristic-driven stochastic search
based system and currently the state of the art for optimal synthesis
of RISC-V programs~\cite{stoke}. To accommodate the simplifications we
made to the RISC-V language, we re-implement Stoke's MCMC
search to allow it to search over the RISC-V space and reward function that our
method uses. We refer to this baseline as "MCMC." We selected a value for the MCMC $\beta$ hyperparameter by testing on a holdout validation program, and found that $\beta = 2$ works well.

\subsection{Program Length}
\label{subsec:program_length}

Because program length is a major determinant of search space
complexity, we characterized each method's behavior as a function of
the length of ground truth program. Figure \ref{fig:prog_length} shows
the results on our synthetic benchmark for lengths between 3 and 10
lines. RLGTS solves at least twice as many programs as the best baseline, the bandit. The addition of the priority search tree improves our performance over the Q-function-only ablation by 70-100\% consistently.

A fraction of the programs generated have a convex reward function space and can be trivially solved by convex descent using a best-first search algorithm. Specifically, for a program to be considered convex, among the rewards of all possible length 1 programs the first line of a satisfying program ranks highest, among all length 2 extensions of that best length 1 program a length 2 subprogram of a satisfying program is the highest ranked, and so on until a satisfying program is found. We found in testing that about 30\% of programs we generate have this property. As these cases are easily solvable via a na\"ive best-first search algorithm and are expected to be rare among programs of human interest, we filter and remove them from the remainder of our evaluation. Success rates on convex programs are included separately in figure \ref{fig:prog_length}, where we observe that RLGTS solves more than 90\% of such cases, albeit at the cost of more computation than best-first search.

\textbf{Search Depth Limit}

\begin{figure}[t]
	\begin{center}
		\includegraphics[width=0.5\linewidth]{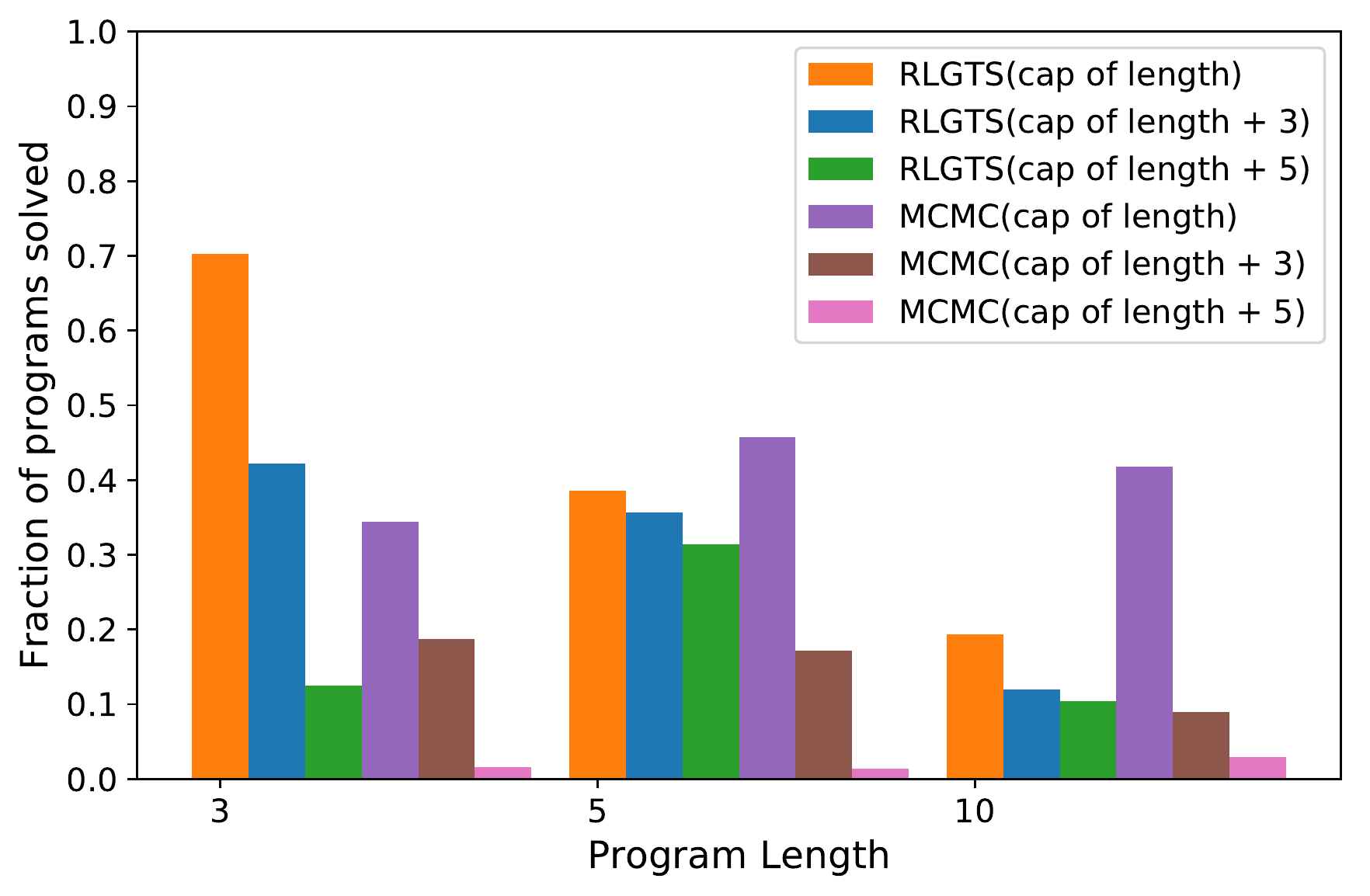}\includegraphics[width=0.5\linewidth]{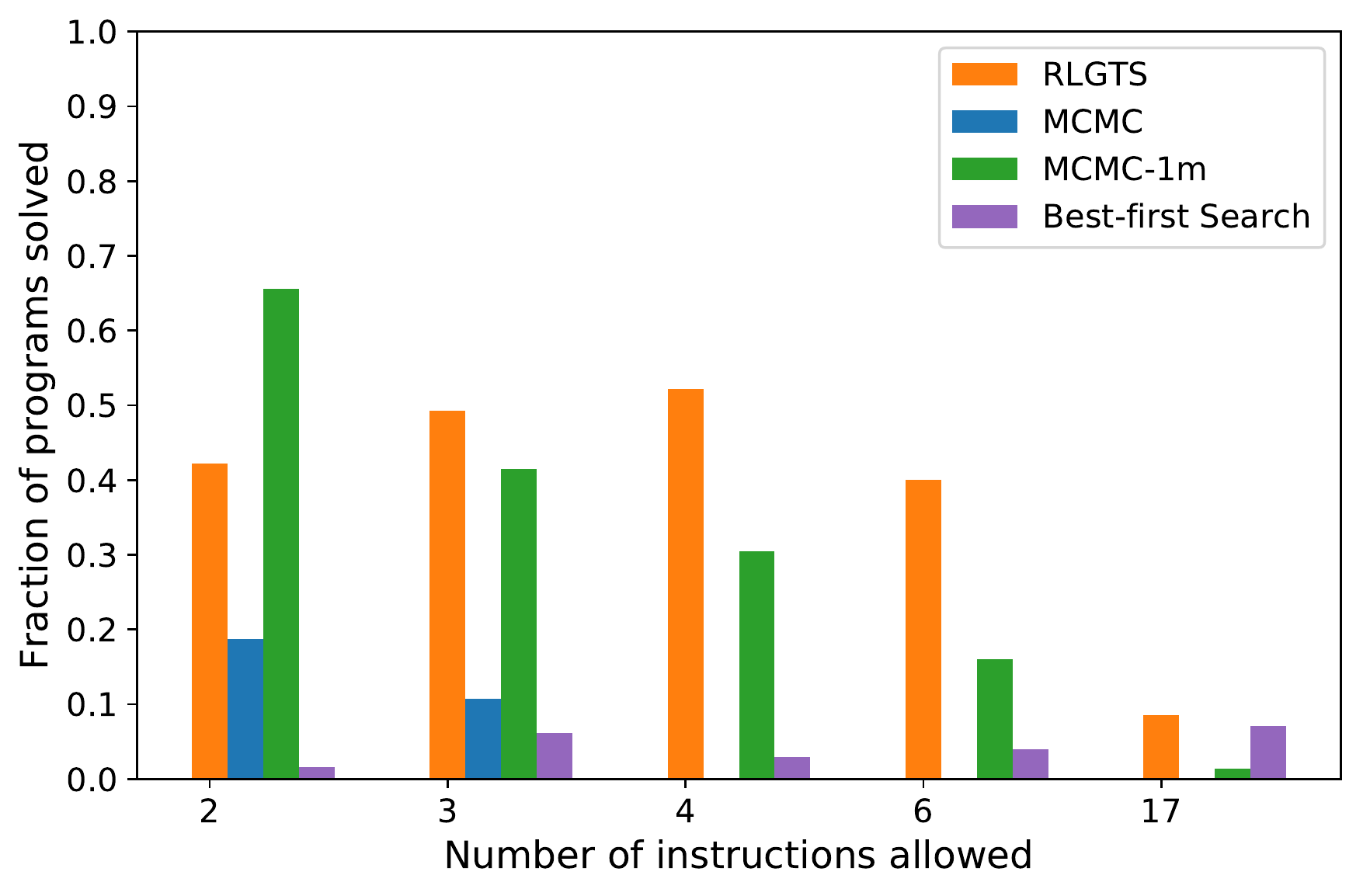}
	\end{center}
    \caption{Comparison of performance between RLGTS and MCMC as a function of program length(left) and number of instructions allowed(right). RLGTS retains better performance as the difference between program length and the search cap increases. Surprisingly, MCMC performance decreases only slightly if at all as program length increases.  MCMC sample efficiency drops off much more rapidly as the number of instructions searched over increases. "MCMC-1m" shows the performance of MCMC when its timeout was increased to 1,000,000 attempts versus the 20,000 attempt limit otherwise used. 
    }
    \label{fig:search_depth}
\end{figure}

While RLGTS readily outstrips the na\"ive search and multi-armed bandit baselines for various lengths and search depths, we found during testing that MCMC search performance on this benchmark is highly dependent on the difference between the true program length and the search depth limit on maximum program length we place. This is consistent with its original proposed use-case of super-optimization, wherein an existing program is provided as part of the specification, which allows for a good bound on search depth to be defined~\cite{stoke}. Because of this sensitivity, we compare RLGTS against MCMC for fixed differences between ground-truth program length and maximum search depth,
We show the results of this comparison in figure \ref{fig:search_depth}. While MCMC performs competitively with RLGTS on longer programs when the target program length is known, its performance degrades rapidly when the maximum program length diverges from it, losing 50-80\% of its performance when the cap is 3 lines above the GT length, and is effectively 0 at length + 5. While RLGTS is also adversely affected by not knowing the GT program length, the impact is smaller, in the range of 10-50\%.

\subsection{Action Space Complexity}
\label{subsec:action_space}

In addition to program length, we also explore the number of
single-line programs expressible as a factor for performance. 
Figure \ref{fig:search_depth} shows performance for RLGTS and MCMC, as well as a best-first search baseline, as a function
of the number of instructions allowed, ranging from 2 instructions to all 17
instructions defined in Figure~\ref{fig:lang}. All programs are of length 3, with a maximum search depth of 6. Interestingly, we note that RLGTS retains performance of between 40-50\% for small numbers of instructions,
while MCMC performance decreases much more rapidly and cannot solve any programs allowing 4 or more instructions within 20,000 attempts. To get an estimation of how many attempts MCMC requires to solve programs with more instructions, we ran MCMC with an attempt limit of 1 million. With an attempt limit 50 times higher than RLGTS, MCMC is able to exceed our performance on programs of 2 instructions and match it on 3. MCMC is still unable to solve any programs allowing all 17 instructions, however, while RLGTS solves around 9\% in far fewer attempts.

\section{Discussion}

Here, we have presented reinforcement learning guided tree search, a new approach for
program synthesis powered by reinforcement learning. This approach is
general and flexible, with up to 400\% better performance than the state of the art in 
traditional search-based methods on cases where search achieves a non-trivial success rate.
It's performance suggests that with further research RLGTS may be able to scale to solve more complex programs that cannot be solved by previous methods.

\subsubsection*{Acknowledgments}
We would like to thank Kyle Genova, David Walker, Olga Russakovsky, and Aarti Gupta for useful discussions and feedback. We would also like to thank Maciej Balog for information and advice relating to the Deepcoder language and system.

\bibliography{local}

\end{document}